%% file: 0-main.tex
\documentclass[runningheads]{llncs}
\usepackage{graphicx}
\usepackage{subcaption}
\usepackage{multirow}
\usepackage{tablefootnote}
\usepackage{graphicx}
\usepackage{tabularray}
\usepackage{svg}
\usepackage{multirow}
\usepackage{array}
\usepackage{tabularx} 
\usepackage{amsfonts}
\usepackage{booktabs}
\usepackage{siunitx}
\usepackage{makecell}
\usepackage[compress]{cite}
\usepackage{diagbox}
\usepackage[skip=2pt]{caption}
\usepackage{subcaption}
\usepackage{makecell}
\usepackage{siunitx}
\usepackage{array}
\usepackage{algorithm}
\usepackage{algpseudocode}
\usepackage{xcolor}
\usepackage{amsmath} 
\usepackage{amsfonts} 
\usepackage{amssymb} 
\usepackage{mathrsfs} 
\usepackage{booktabs} 
\usepackage{diagbox} 

\newcommand{\eg}{{\it e.g., }}
\newcommand{\ie}{{\it i.e., }}

\def\BibTeX{{\rm B\kern-.05em{\sc i\kern-.025em b}\kern-.08em
    T\kern-.1667em\lower.7ex\hbox{E}\kern-.125emX}}
\makeatletter
\@addtoreset{equation}{section}
\makeatother

\makeatletter
\def\@copyrightspace{\relax}
\makeatother

\begin{document}
\title{Unveiling the Unseen: Exploring Whitebox Membership Inference through the Lens of Explainability}
\author{Chenxi Li \and
Abhinav Kumar \and
Zhen Guo \and
Jie Hou \and
Reza Tourani}
\authorrunning{C. Li et al.}
\titlerunning{Unveiling the Unseen: MIA Explainability}
%
\institute{Saint Louis University, St. Louis MO 63103, USA\\
\email{\{chenxi.li,abhinav.kumar,zhen.guo.2,jie.hou,reza.tourani\}@slu.edu}}

\maketitle 
\vspace{-0.15in}
\begin{abstract}
The increasing prominence of deep learning applications and reliance on personalized data underscore the urgent need to address privacy vulnerabilities, particularly Membership Inference Attacks (MIAs). Despite numerous MIA studies, significant knowledge gaps persist, particularly regarding the impact of hidden features (in isolation) on attack efficacy and insufficient justification for the root causes of attacks based on raw data features.
In this paper, we aim to address these knowledge gaps by first exploring statistical approaches to identify the most informative neurons and quantifying the significance of the hidden activations from the selected neurons on attack accuracy, in isolation and combination. Additionally, we propose an attack-driven explainable framework by integrating the target and attack models to identify the most influential features of raw data that lead to successful membership inference attacks. Our proposed MIA shows an improvement of up to 26\% on state-of-the-art MIA.

\end{abstract}
%
%
%

\input{1-Introduction}

%
\input{2-RelatedWork}
%
\input{3-ThreatModel}

%
\input{4-Methodology}
\input{5-XAI}
%
\input{6-Experiments}

%

\section{Conclusion}
In this paper, we conducted an in-depth analysis of hidden activations and their impact on the efficacy of whitebox MIAs. We proposed a novel membership-guided neuron selection mechanism that analyzes the distributions of member and non-member data samples to identify the most influential subset of neurons for the attack. Employing explainable AI techniques, we conducted feature importance analysis to inform an ensemble attack framework, resulting in a 26\% improvement in MIA accuracy over the state-of-the-art. Additionally, we introduced an explainable attack-driven framework, by integrating the target and attack models, to identify the impact of raw data features on our MIA's success. Our analysis revealed that the intersection between raw data features significant for the target classification task and those important for the MIA is capped at an SSIM score of 0.42. Leveraging insights from our analysis, we plan to develop a defensive measure against MIAs using adversarial samples by perturbing significant raw data features crucial for MIA success but not essential for classification, enhancing attack resiliency without compromising classification performance.

\bibliographystyle{splncs04}
\bibliography{0-main}

\input{7-Appendix}

\end{document}

%% file: 1-Introduction.tex
\section{Introduction}
\label{sec:introduction}
\begin{figure}[t]
    \centering
    \begin{subfigure}[t]{0.33\textwidth}
        \includegraphics[width=\textwidth]{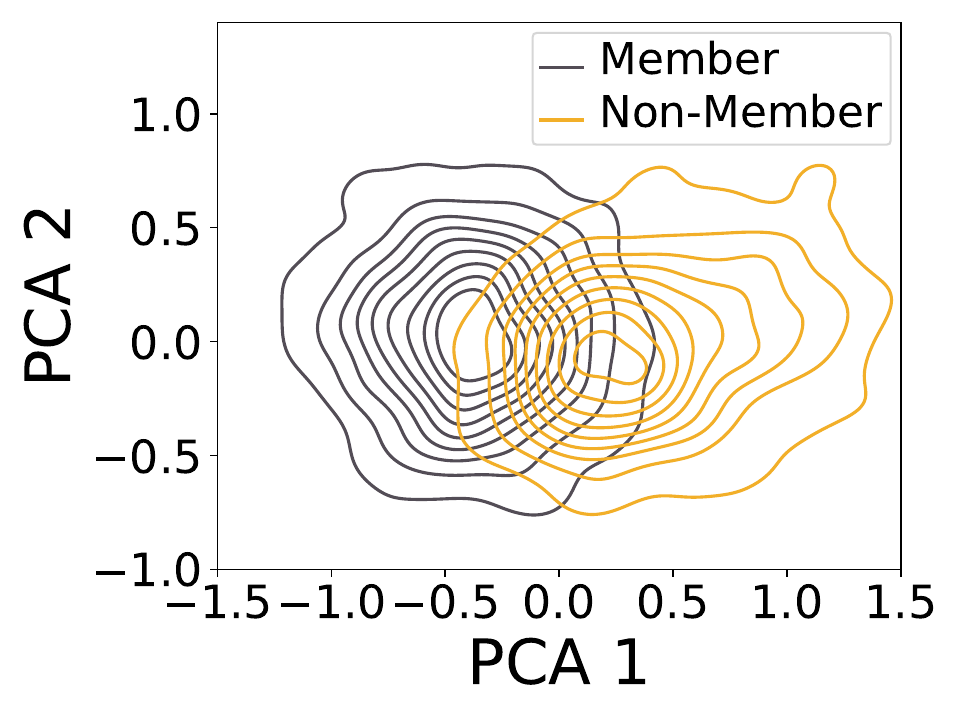}
        \vspace{-0.2in}\caption{100\% Activations}
    \end{subfigure}%
    \hfill
    \begin{subfigure}[t]{0.33\textwidth}
        \includegraphics[width=\textwidth]{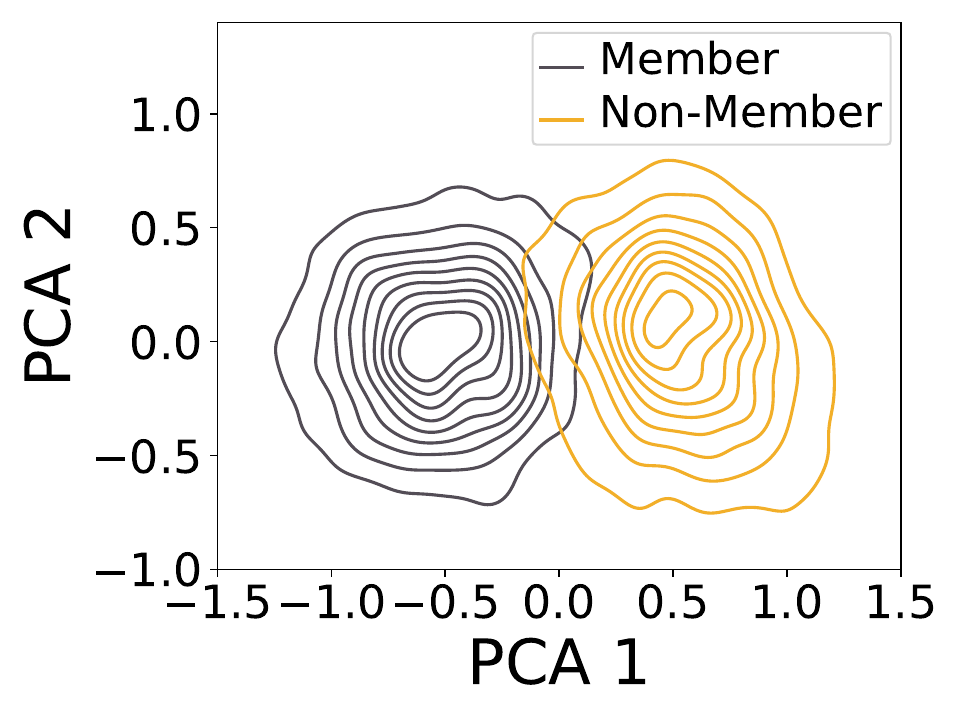}
        \vspace{-0.2in}\caption{60\% Activations}
    \end{subfigure}%
    \hfill
    \begin{subfigure}[t]{0.33\textwidth}
        \includegraphics[width=\textwidth]{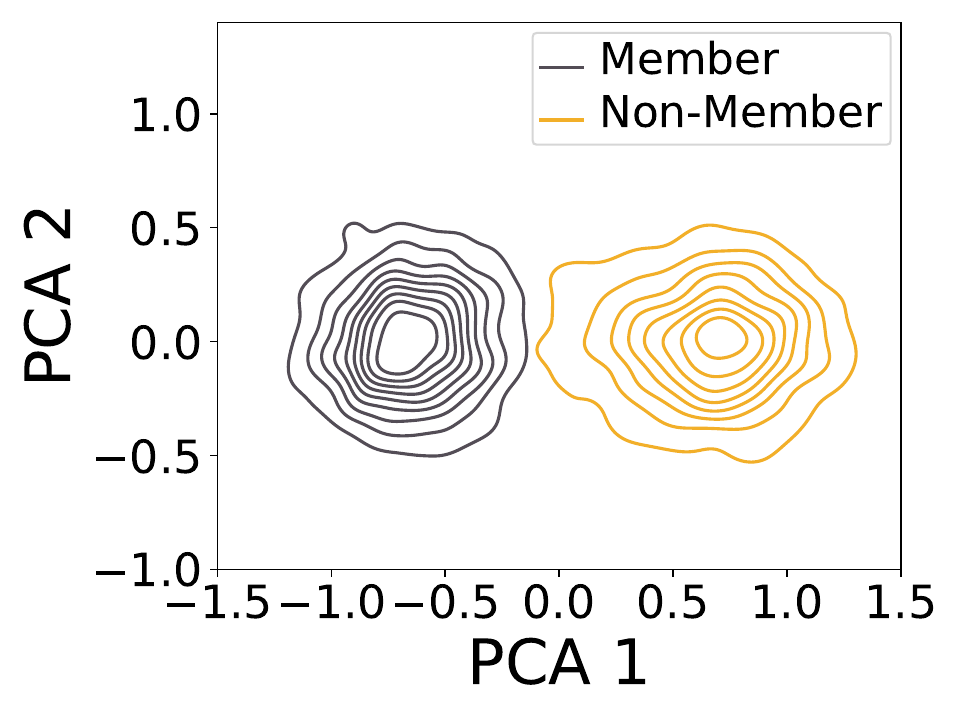}
        \vspace{-0.2in}\caption{20\% Activations}
    \end{subfigure}
    \caption{\small Performing PCA analysis on second-to-last layer's output from both member and non-member data samples using a ResNet architecture trained on the UTKFace dataset. The results reveal that selecting the top 20\% of the most influential neurons results in a more pronounced separation between member and non-member samples compared to using all neurons.}
    \label{fig:PCA}
    \vspace{-0.25in}
\end{figure}

%
The recent developments in machine learning (ML) have led to the development of innovative solutions in various domains, ranging from medical imaging and precision medicine to impressive deep learning-based text to image/video generation. The success of such data-driven applications relies heavily on the availability of high-quality data, which often includes sensitive or private information. However, the use of personal data in large deep learning model training has raised serious security and privacy concerns due to the potential privacy attacks on machine learning models, particularly membership inference attacks (MIAs)~\cite{ShoStrSon16,long2018understanding,yeom2017,homer2008resolving,nasr2018comprehensive,rezaei2021difficulty,Liu_Wu_Yu_Zhang}. The MIAs allow the adversaries to infer private information from a target ML model, \eg membership of a single data sample in the training set. The state-of-the-art whitebox MIAs, where adversaries have complete access to the target model, have identified significant attack features, including gradients, classification loss, prediction posteriors, labels, and hidden layers embedding~\cite{nasr2018comprehensive,liu2021ml}.
We argue that choosing the most influential hidden activations enhances the robustness and efficiency of MIA, compared to selecting all hidden activations or their embeddings. Our intuition originates from the notion that only a subset of neurons significantly impacts the target classification task~\cite{bau2020understanding}. To investigate this argument, we conducted a Principal Component Analysis (PCA) analysis to visualize the distributions of \emph{member} and \emph{non-member} samples based on various activation proportions extracted from the last layer before the classification layer of a ResNet18 model trained on the UTKFace dataset (Figure~\ref{fig:PCA}). Notably, we observed a diminishing overlap between the member and non-member distributions, when selecting only a small but influential set of neurons.

In this paper, motivated by this observation, we aim to conduct an in-depth study of the white-box membership inference attack and its root causes using explainable AI techniques. We begin by identifying the most significant neurons that would potentially differentiate the membership data and non-membership data using common statistical methods and then quantitatively analyze their impact on the attack's efficacy. Subsequently, we perform a complementary analysis to unveil the causal relationship between the raw data features, influenced by the identified hidden neurons' features, and the attack's success. Through this study, we aim to answer two key research questions:
%
\textit{(i) Which hidden activations exhibit greater significance in the context of MIAs and to what extent?} To answer this question, we introduce a novel white-box MIA framework, which selects the most influential hidden activations across all layers using various statistical methods, such as Kullback-Leibler divergence and random forest. Subsequently, we conduct a series of MIAs using these identified features in isolation and utilize SHapley Additive exPlanations (SHAP)~\cite{LundbergLee2017} to quantify these features' significance and explain their directional influence. 
To overcome the limitations of individual attacks, we leverage the insights from our analysis and adopt ensemble learning to combine the most effective attack strategies. 
\textit{(ii) How can we pinpoint the raw data features most influential in the attack's success, considering the interplay between the input and the identified hidden activations?} 
To answer this question, we propose an attack-driven explainable framework by combining the target and attack model into an integrative module, seamlessly explaining the input and attack features that contribute to membership inference attacks. The proposed attack-driven framework is fully differentiable among raw input data, target model, and attack inference. Given the potential disparity between the two neuron sets critical for the target classification task and those significant for the MIA, respectively, we perform an empirical quantitative analysis on the intersection between these two neuron sets. This analysis facilitates the development of tailored defensive strategies against MIAs, ensuring robust protection without compromising target classification performance.


In summary the {\bf novel contributions} include:

\noindent $\bullet$ Rigorous investigation into the utility of hidden activations, involving the selection of different subsets from different layers of a neural network using statistical techniques, in identifying membership data samples. Our analysis shows that a small subset of neurons has the most influence on the MIA efficacy. 

\noindent $\bullet$ Design of a novel explainable framework to identify and reason the root causes of the MIA success concerning the raw data features. The design of this attack-driven framework is generic, which makes it applicable to other privacy attacks, \eg property inference and data reconstruction.

\noindent $\bullet$ Conducting extensive experiments to assess the efficacy of the proposed MIA on major benchmark datasets of varying complexity using two architecturally different deep neural networks. The experimental results demonstrate that our MIA improves the results reported in~\cite{liu2021ml} across three benchmark datasets -- 3\%--25\% for FMNIST, 5\%--26\% for UTKFace, and 3\%--16\% for STL10.




%% file: 2-RelatedWork.tex
\section{Related Work}
\label{sec:relatedwork}

The increased utilization of sensitive data in the training of machine learning models has led researchers to extensively explore inference attacks targeting the sensitive training samples~\cite{ateniese2015hacking,fredrikson2014privacy,tramer2016stealing, dougherty2023stealthy}. One of the most prominent attacks is the Membership Inference Attack (MIA) where the adversary, during the inference phase, aims to infer whether a given sample was utilized in training~\cite{ShoStrSon16}. To conduct these attacks the adversary often relies on an auxiliary dataset, also referred to as a shadow distribution, which is similar to the training distribution and can potentially share some samples with the training distribution~\cite{liu2021ml}. Based on the threat model and attack features, the attack methodologies can broadly be classified into black-box and white-box attacks.
In a {\bf black-box} setting the adversary only has an API-level access to the ML model and relies on the prediction vectors to conduct the attack~\cite{ShoStrSon16,long2018understanding,yeom2017,homer2008resolving}.
{\bf White-box} settings assume a stronger adversary who also has access to the model parameters, activations, and gradients~\cite{nasr2018comprehensive,rezaei2021difficulty,Liu_Wu_Yu_Zhang}.
Even though in the white-box setting the adversary has complete access to the target model, researchers have only seen a marginal improvement over black-box attacks~\cite{nasr2018comprehensive, liu2021ml}.

While the prior work has put forward several hypotheses for the success of MIA, it fails at explaining the behaviors we notice in the white-box setting. This makes explainable ML solutions an ideal candidate as they have also been previously used to assess the security and privacy of ML models~\cite{hu2022membership, ye2022enhanced, Liu_Wu_Yu_Zhang}. However, the explainability of white-box attacks is still underexplored.

In this work, we explore the limitations of existing state-of-the-art white-box attacks and develop a neuron selection framework, leading to a stronger white-box attack. We also utilize explainable ML techniques to develop a framework to identify and evaluate the most important attack features.


%% file: 3-ThreatModel.tex
\section{Threat Model and Security Assumptions}
\label{sec:threat}

We consider an honest-but-curious adversary, who attempts to infer information about the membership status of a data sample in the training process, without sabotaging the training process. The adversary constructs the attack without access to the training data.

\noindent
\textit{Target \& Shadow Models}: To investigate how the activation outputs from different proportions of hidden neurons affect the privacy attacks, we consider a white-box MIA, where the adversary possesses complete knowledge of the target model, denoted as \(M^T(.)\), including the model's architecture and its pre-trained parameters. This white-box setting is relevant in cases where the model's parameters are exposed through reverse engineering~\cite{oh2019towards}. We also define a shadow model, denoted as \(M^S(.)\), that allows the adversary to reproduce a model with similar performance as the target model by training on an independent dataset, specifically adopted for developing the attack models.

\noindent
\textit{Target \& Shadow Datasets}: In our study, the target model (\(M^T(.)\)) and the shadow model (\(M^S(.)\)) are trained on two independent datasets, referring to as the target dataset ($\mathcal{D}^{T}$) and the shadow dataset ($\mathcal{D}^{S}$), respectively. We consider a scenario where the adversary has little knowledge of the private training dataset ($\mathcal{D}^{T}$), however, the adversary has the access to a public shadow dataset ($\mathcal{D}^{S}$), which is assumed to follow the same underlying distribution as $\mathcal{D}^{T}$ but has no overlapping data points ($\mathcal{D}^{S} \cap \mathcal{D}^{T} = \emptyset$). 
There are various approaches for building shadow datasets~\cite{ShoStrSon16} and the adversary's access to the shadow dataset is a common assumption~\cite{ShoStrSon16,nasr2018comprehensive,dougherty2023stealthy}. 

\noindent
\textit{Attack Model}: Our research utilizes the shadow dataset ($\mathcal{D}^{S}$) and the shadow model (\(M^S(.)\)) to create the attack feature datasets for training the white-box MIA model. The attack model is then used to determine the membership status of data samples in the target training dataset ($\mathcal{D}^{T}$), using the attack features derived from the target model (\(M^T(.)\)).

%% file: 4-Methodology.tex
\section{White-box Attack Methodology}
\label{sec:methodology}
\begin{figure}[t]
    \centering
    \includegraphics[width=\textwidth]{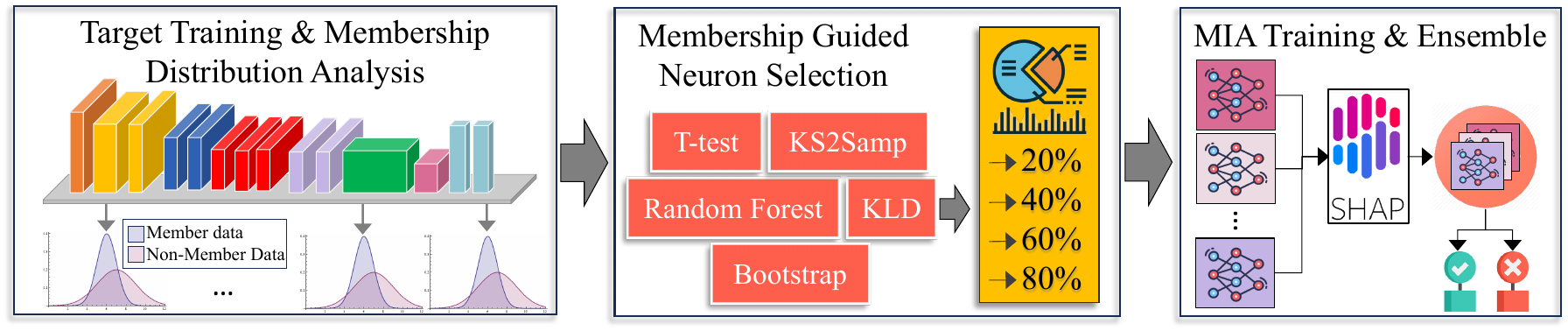} 
    \caption{\small Our proposed MIA framework analyzes membership distribution of data samples for selecting the most influential subset of neurons as attack features. For ensembling the final MIA model, it uses SHAP to select the most significant MIA models.}
    \label{fig:pipe}
    \vspace{-0.2in}
\end{figure}

Our white-box MIA framework is implemented via a three-step pipeline: \emph{(1) Target Model Training and Membership Distribution Analysis}, \emph{(2) Membership Guided Neuron Selection}, and \emph{(3) MIA Training \& Ensemble}, which is illustrated in Figure~\ref{fig:pipe}.
This framework enables a comprehensive analysis of the diverse model attributes, particularly the activation outputs of hidden layers, that potentially causes the privacy leakages of the training data. We employ five statistical approaches to perform the comparative analysis of the distribution of hidden neurons' output and distinguish between the training data (\ie membership data) and test dataset (\ie non-membership data). This proposed membership distribution analysis will identify the critical hidden neurons that exhibit significant differences in predictive patterns between the membership and non-membership datasets within the target model (Figure \ref{fig:PCA}). The statistical comparison also guides the selection of neurons of models, specific to feature patterns in membership data, enhancing the effectiveness of attack models. In the following subsections, we elaborate on each step of this framework. 

\subsection{Target Model Training and Membership Distribution Analysis}
%

In this work, we utilize a pre-defined deep learning architecture to train the target model on a image dataset $D^T = \{(x_i, y_i) \mid i \in \{1, \cdots, N\}, y_i \in \{1, \cdots, K\}\}$, where $y_i$ represents one of the $K$ distinct classes for each data point in the target dataset. The target model (\(M^T(.)\)) is trained to achieve high classification performance on the dataset, adhering to standard deep neural network training configurations, including the optimization of cross-entropy loss function using stochastic gradient descent algorithm. Similarly, we train the shadow model (\(M^S(.)\)) on the shadow dataset ($D^S$) with a similar underlying distribution as the target dataset ($D^T$), to achieve similar performance as the target model.

To investigate patterns specific to membership and non-membership data, we extract the intermediate activation values from layer $h$ of the shadow model (\(M^S(.)\)) for the membership data, denoted as $[a|D_{\text{mem}}]_{h}$, and non-membership data, denoted as $[a|D_{\text{non-mem}}]_{h}$, respectively. We generalize this notation to $[a|\tilde{D}]_M = \{[a]_1, [a]_2, \cdots, [a]_H\}$ for all the $H$ layers of model (\(M^S(.)\)) across the dataset $\tilde{D}$, where $\tilde{D} \in \{D_{\text{mem}}, D_{\text{non-mem}}\}$.  As illustrated in Figure \ref{fig:pipe}, we compare and analyze the distributions of activation patterns for both membership data ($D_{\text{mem}}$) and non-membership data ($D_{\text{non-mem}}$) at each layer of shadow model $M_S(.)$ to identify the key neurons that significantly distinguish between member and non-member data points.
The membership distribution analysis provide evaluation of neuron significance across different layers and dataset, and further enable the neuron selection in the subsequent step to design better attack features to enhance the effectiveness of attack models.

\subsection{Membership Guided Neuron Selection}
\label{sec:neuronselection}

To perform the statistical comparison between the activation patterns of neurons from datasets $D_{\text{mem}}$ and $D_{\text{non-mem}}$ within a specific neural network layer, we employ the statistical methods on intermediate activation values at each layer of: $[a|D_{\text{mem}}]_M$ and $[a|D_{\text{non-mem}}]_M$. 
First, we conducted an independent two-sample t-test for each neuron's activation across the two datasets, $[a_j|D_{\text{mem}}]_{M,i}$ and $[a_j|D_{\text{non-mem}}]_{M,i}$, where $i$ denotes the i-th layer and $j$ represents the j-th neuron within layer $i$ of the model $M$ \cite{manfei2017differences}. The resulting p-values from these tests for all neurons within each layer were used to determine statistically significant differences in activation patterns between membership data and non-membership data. Neurons with p-values $<0.05$ within the same layer were ranked in ascending order of their p-values to identify the most significant neurons associate with membership data. Second, we applied the two-sample Kolmogorov-Smirnov test (KS2Samp) to compare the underlying continuous distributions of activation patterns between the membership data and non-membership data for every neuron within each layer \cite{berger2014kolmogorov}, followed by the neuron ranking according to their p-values. Further, we employed the Kullback–Leibler divergence (KLD) method to quantify differences between the two probability distributions of activation patterns for membership data and non-membership data \cite{joyce2011kullback}. Different from the t-test, we rank the neurons in descending order, with higher KL-divergence values indicating greater divergence between the two distributions. The forth statistical method employed the bootstrapping algorithm, which resamples activation values with replacement for each neuron, to assess differences in the central tendency of activation values between the membership data and non-membership data \cite{singh2008bootstrap}. The neurons were then ranked based on their magnitude of bootstrapped differences. lastly, the fifth method adopts the machine learning based approach, Random Forest, to identify the most important neurons that discriminate between the membership data and non-membership dataset \cite{liu2012new}. The feature importance analysis was performed for each neuron within same layer to assign the feature important scores for every neuron, where higher scores suggest most significant neurons with activation patterns specific to membership data. The neurons were then ranked based on their importance scores.

To delve deeper into the impact of neurons in each network layer on the success of MIAs, we select the top 20\%, 40\%, 60\%, and 80\% of neurons in each layer based on their scores to explore the significance of different proportions of neurons on the effectiveness of MIAs. We formalize the attack dataset, $\mathcal{D_S}^{Attack}$, from the shadow dataset $(D_S)$, as follows: 
%
\begin{align*}
\mathcal{D}^S_{\text{Attack}} &= \left\{(\mathbf{[I]}, y) : \mathbf{[I]} = \left([a]_h^{\mathcal{T}}(\bar{x}), \mathbf{p}(\bar{x}), \hat{y}(\bar{x}), \mathcal{L}(\bar{x}), \nabla \mathcal{L}(\bar{x})\right), \forall \bar{x} \in \mathcal{D}^{S}\right\} 
\end{align*}
The dataset ($\mathcal{D}^S_{\text{Attack}}$) comprises of binary membership labels \( y \in \{0,1\} \), and diverse attack feature sets, denoted as $[I]$, each representing a type of model information extracted from the shadow model ($M_S(.)$). Specifically, we consider the five types of attack features to infer the private information of the training data using MIAs. First, $[a]_h^{\mathcal{T}}$ represents the intermediate activation values from layer $h$ where \( \mathcal{T} \) denotes the neuron selection thresholds of \{20\%, 40\%, 60\%, 80\%\}. The thresholds are used to select portions of the neurons in the hidden layer to generate activation values as the input features in attack models. We also include the vector of output probabilities, denoted as $\mathbf{p}(\bar{x}) = \mathcal{M}(\bar{x})$, the predicted class label, determined as \(\hat{y}(\bar{x})\) $= \operatorname*{argmax}_c P(y=c|\bar{x})$, and the predictive loss, represented as $\mathcal{L}(\bar{x}) = \mathcal{L}(\hat{y},y)$. Following the work in MIA studies~\cite{nasr2018comprehensive,liu2021ml}, the gradients of predictive loss ($\mathcal{L}$) with respect to set of model parameters ($[w]$), denoted as $\nabla \mathcal{L}(\bar{x}) = [\dfrac{\partial \mathcal{L}}{\partial w_1}, \dfrac{\partial \mathcal{L}}{\partial w_2},\ldots,\dfrac{\partial \mathcal{L}}{\partial w_m}]_{x=\bar{x}}$, are included in the attack feature sets. In this work, we investigate the optimal neuron selection thresholds (\( \mathcal{T} \)) using membership inference attacks that utilize the attack features, which include the activation values of the selected neurons.  
\vspace{-0.1in}
\subsection{MIA Training and Ensemble}
\vspace{-0.0in}

Following the neuron selection threshold and attack features defined in Section \ref{sec:neuronselection}, we feed both membership data and non-membership data from the shadow dataset into the shadow model $M^S(.)$. We then employ the membership-guided neuron selection method to identify subsets of neurons (\{20\%, 40\%, 60\%, 80\%, 100\%\} in our experiments) using five statistical methods across all intermediate layers of the shadow model. Each selection threshold ($\mathcal{T}$) for neurons in each layer, guided by each method of membership distribution analysis, generate distinct attack feature datasets used for training the MIAs models. The training of MIAs model follows the similar protocols introduced in~\cite{liu2021ml}, in which attack features from each neuron selection threshold and statistical method are fed into different 2-layer neural network models to generate the attack embeddings. These embeddings are then concatenated and fed into a 4-layer neural network model for membership inference. In total, we trained twenty separate MIA models, using attack features defined by four neuron thresholds \{20\%, 40\%, 60\%, 80\%\} for each of the five statistical methods. We then evaluated the accuracy of each MIA model on both the shadow dataset ($D^S$) and the target dataset ($D^T$).

To further increase the accuracy of MIAs, we also investigated the effectiveness of ensembling techniques that combine the prediction from individual MIA models. Instead of simply combining all twenty MIA models using standard ensembling, we implemented an explainable-AI guided strategy for selective model combination. We applied SHAP analysis to the predictive probabilities of all MIA models and quantified their contributions to membership inference. The explainable analysis was conducted on shadow dataset using MIA probabilities derived from both shadow and attack models. SHAP analysis determines the importance score to each of input MIA models, enabling the strategic selection of the top \( k \) models for ensemble integration. Our ensemble framework utilizes a stacking methodology, incorporating a meta-model and three base models: decision tree, random forest, and support vector classifier. These base models provide diverse predictive insights from the given attack probabilities of MIA models. The predictive results from these base models are then fed into the meta-model, which utilizes logistic regression, for the final membership inference attack. 

%% file: 5-XAI.tex
\section{Explainable Membership Inference Methodology}
\label{sec:XAI}
\begin{figure}[t]
    \centering
    \includegraphics[width=0.9\textwidth]{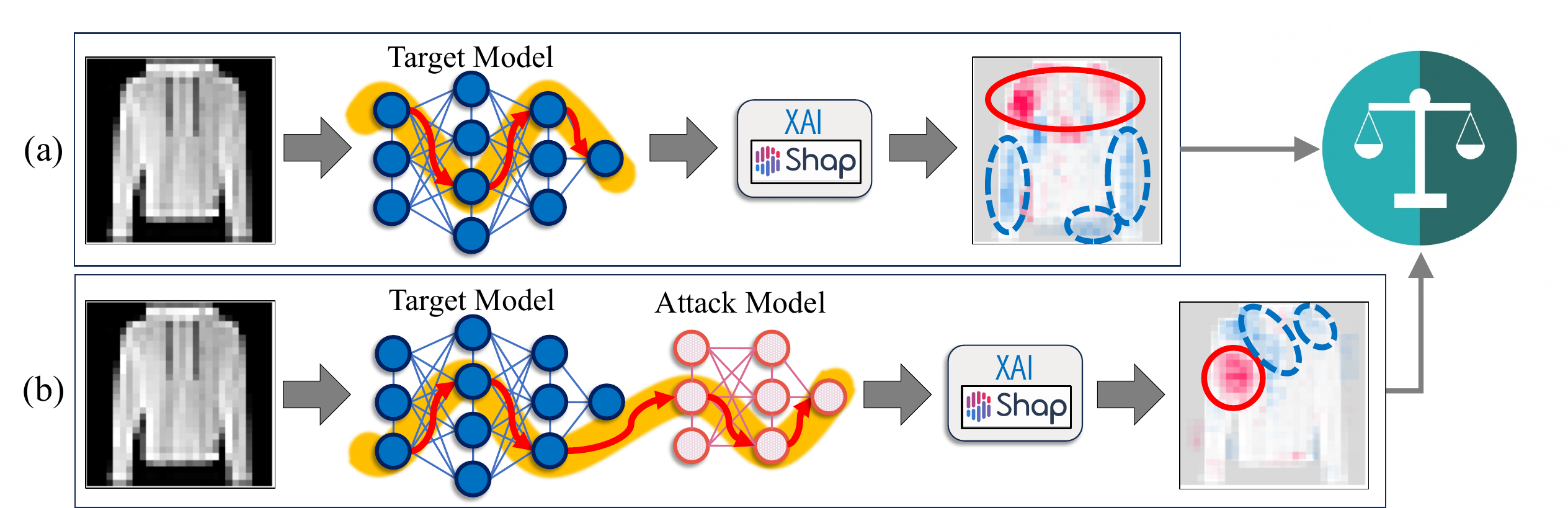} 
    \caption{\small The proposed explainable MIA framework, in which the raw feature importance will be selected based on the target model (a), as well as the combination of the target and attack models (b), to identify the significant raw features that leak more information regarding sample membership.}
    \label{fig:XAI-Framework}
    \vspace{-0.3in}
\end{figure}

Explainable AI (XAI) techniques have been used to explain various facets of machine learning models, such as predictions, robustness, and fairness. However, there remains a lack of attention in applying XAI to understand the machine learning privacy implications, particularly in the whitebox setting. This has led to the success of MIAs being solely attributed to the overfitting rate~\cite{nasr2018comprehensive,liu2021ml}.
To explain the factors contributing to white-box MIAs, we propose an attack-driven explainable framework (Figure~\ref{fig:XAI-Framework}). Our framework identifies two sets of influential features in the raw data: those crucial for the target classification task and those important for the attack's success. It then measures the overlap between these sets. The insights derived from our framework can be used to design customized defenses against membership inference attacks.

%
\noindent {\bf Target Model Interpretable Module.} Aiming to understand how the target model's predictions relate to its input and identify its influential features, we utilize the SHAP mechanism in designing our proposed framework. 
Specifically, our framework uses the DeepExplainer tool to compute the SHAP values on the input data (\eg image). This computation analyzes how variations in the pixel values of the image influence the model's predictions, thereby quantifying each pixel's contribution. The derived SHAP values are then visualized on the input image, with contributions positively affecting the model's predictions highlighted in red and those negatively influencing highlighted in blue. As illustrated in Figure~\ref{fig:XAI-Framework}(a), this visualization highlights the segments of the input data that influence the target model's prediction. We note that the explainability of the target classification task is not our primary focus. However, such an analysis provides a reference for comparison with the explainability analysis of the attack model on the same input sample, enabling the quantification of their intersection.

\noindent {\bf Attack Model Interpretable Module.}
Applying the explainability analysis on the attack model alone will provide some insight into the significance of the attack features (discussed in Section~\ref{sec:neuronselection}). While beneficial, such insights are limited to the processed features, \eg loss value and predicted label, falling short of revealing the influence of the raw data features on the attack's success. The challenge in performing this analysis lies in the fact that the explainable framework requires access to raw data to identify and assess the impact of influential raw features, whereas the attack model only accepts processed features.

To address this challenge, 
we propose a cascaded architecture to integrate the target and attack models into one module (Figure~\ref{fig:XAI-Framework}(b)).
Our proposed architecture includes three components, comprising the target model, the \emph{forward hook} module, and the attack model. In connecting the target and attack models, we integrate a forward hook module, which captures the activations of the target model at specific layers and relay them 
to the attack model for forward propagation. Thus, allowing the 
input data to flow from the target model to the attack model, creating a seamless data-flow. 
%

After integrating the target and attack models, we perform the explainable analysis similar to that of the target model. Specifically, an input image is fed into the combined target-attack model, which leads to an attack prediction, \ie \textit{member} or \textit{non-member}. Utilizing the reverse inference methodology in SHAP mechanism, the framework then analyzes the variations in the pixel values to quantify pixels' contributions to the success of the attack. Subsequently, the importance of features is visualized on the input image, resulting in another highlighted input image that identifies the most critical features specific to MIAs. It also allows the identification of the specific image regions and features that significantly influence the target model's classification decision, as well as those that leak sensitive information regarding the membership status of the data.
Finally, the framework compares the two highlighted images and quantifies their similarities using metrics such as structural similarity index measure \cite{brunet2011mathematical}.


%% file: 6-Experiments.tex
\section{Experiments}
\label{sec:Experiments}
This section elaborates on the experimental setup and evaluation results. Due to space limitations, we provided hyperparameters and detailed architectures in the Appendix (Section~\ref{appendix}).

\subsection{Datasets}
\label{datasets}
We used the following datasets in our evaluation:

\noindent $\bullet$ STL10 dataset includes 10 different classes. Each class consists of 500 training and 800 test images each 96 x 96 pixels. The images are clustered into 10 classes based on the picture in the image~\cite{coates2011analysis}.

\noindent $\bullet$ FMNIST is a fashion image dataset, which consists of 60,000 training images and 10,000 test images each 28 x 28 pixels. The images are categorized into 10 classes based on the type of fashion item they represent~\cite{Han_Rasul_Vollgraf_2017}.

\noindent $\bullet$ UTKFace is a collection of facial images with more than 20,000 samples, annotated with details regarding age, gender, and ethnicity~\cite{zhang2017age}. 
%

\subsection{Models' Specifications}
\label{models}
{\bf Target Model Architecture} In our study, we test the efficacy of our proposed attack on ResNet18~\cite{he2016deep} and AlexNet~\cite{krizhevsky2012imagenet}. We train the models on the previously mentioned datasets (STL10, FMINST, and UTKFace). The network configurations and training procedures are implemented in PyTorch. Each target model undergoes training for 300 epochs using the Adam optimizer with a learning rate of 1e-5 (refer to Table~\ref{tab:hyperparameters} in the Appendix~\ref{appendix} for more details).

\noindent
{\bf Attack Model Architecture} We designed an encoder for each component of our input vector (Intermediate activations, posterior vector, loss, label, and gradients) to generate embeddings which are then concatenated and passed as input to a four-layer classifier.
We trained this attack model for 50 epochs with a mini-batch size of 64 and binary cross-entropy loss function. We use the Adam optimizer with a learning rate of 1e-5 (Table~\ref{tab:Attack} in the Appendix~\ref{appendix}).
%

\noindent
{\bf Ensemble Model Architecture} To enhance the robustness and predictive performance of our whitebox MIA model, we adopted the stacking approach to ensemble the best-performing attack models. In this approach, multiple base models are combined to generate predictions, which are then used as input features for a meta-classifier that produces the final prediction~\cite{ribeiro2020ensemble}. For the base learners of the stacking ensemble, we selected Decision Trees, Random Forests, and Support Vector Classifiers, and used Logistic Regression as the meta-model.

\noindent
{\bf SHAP Model Architecture} For the SHAP explainability analysis we train a binary classifier with five fully connected layers on the same primary task and dataset as our ensemble meta classifier. We train the model for 10 epochs with a mini-batch size of 32 and a binary cross-entropy loss function. We use the Adam optimizer with a learning rate of 0.001.

\subsection{Evaluation Metrics}
\label{evaluationmetrics}
We evaluate the efficacy of our binary attack model using accuracy and F1-score, where the two classes are member data and non-member data. The test set is divided evenly between both of the classes. We utilize SHAP values to measure and rank the importance of the features selected for our ensemble model which in turn allows us to develop a more efficient white-box attack. We further use the SHAP values to investigate the similarity between the target model's input features impacting the prediction and the input features leading to a successful attack. Finally, we quantify the similarity between the input features by using the structure similarity index (SSIM)~\cite{Wang_Bovik_Sheikh_Simoncelli_2004}.

%

\begin{table}[t]
\centering
\caption{\small Training/Testing accuracy of AlexNet and ResNet18 across datasets.}
\label{tab:accuracy}
\begin{tabularx}{\textwidth}{>{\centering\arraybackslash}p{2.5cm} >{\centering\arraybackslash}p{3cm} *{3}{>{\centering\arraybackslash}X}}
\toprule
\multirow{2}{*}{\makecell{Architecture}} & \multirow{2}{*}{\makecell{Approach}} & \multicolumn{3}{c}{Dataset} \\
\cmidrule(lr){3-5}
& & FMNIST & UTKFace & STL10 \\
\midrule
\multirow{2}{*}{AlexNet} & Ours & 1.000/0.893 & 1.000/0.798 & 1.000/0.543 \\
\cmidrule{2-5}
& ML Doctor~\cite{liu2021ml} & 1.000/0.884 & 1.000/0.792 & 1.000/0.522 \\
\midrule
\multirow{2}{*}{ResNet18} & Ours & 1.000/0.927 & 1.000/0.876 & 1.000/0.565 \\
\cmidrule{2-5}
& ML Doctor~\cite{liu2021ml} & 1.000/0.909 & 1.000/0.852 & 1.000/0.524 \\
\bottomrule
\end{tabularx}
\vspace{-0.2in}
\end{table}

\subsection{Evaluation Results}
\subsubsection{Target Model Performance} Table~\ref{tab:accuracy} shows the accuracy of the two target architectures trained across all the three datasets.
all the models are trained to a hundred percent training accuracy, which is a common practice in MIA literature. When compared to ML Doctor, our target models show a lower overfitting rate, \ie the difference between training and test accuracy, across all dataset and model combinations. As the adversary uses the same architecture and a shadow dataset of the same domain, the shadow models have identical accuracy.

%

\subsubsection{Attack Performance}
\begin{figure}[t]
    \centering
    \begin{minipage}[t]{0.46\textwidth}
        \centering
        \includegraphics[width=\textwidth]{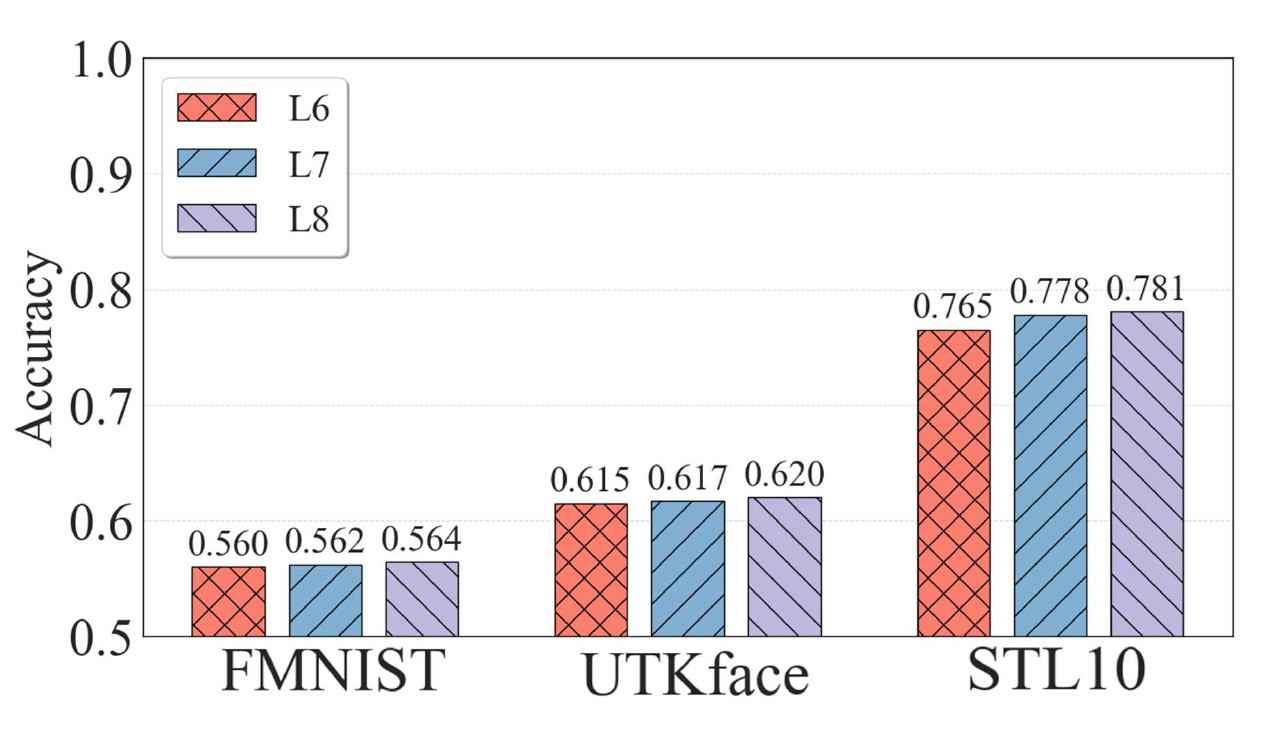}
        {AlexNet}
    \end{minipage}%
    \hspace{0.1in}
    \begin{minipage}[t]{0.46\textwidth}
        \centering
        \includegraphics[width=\textwidth]{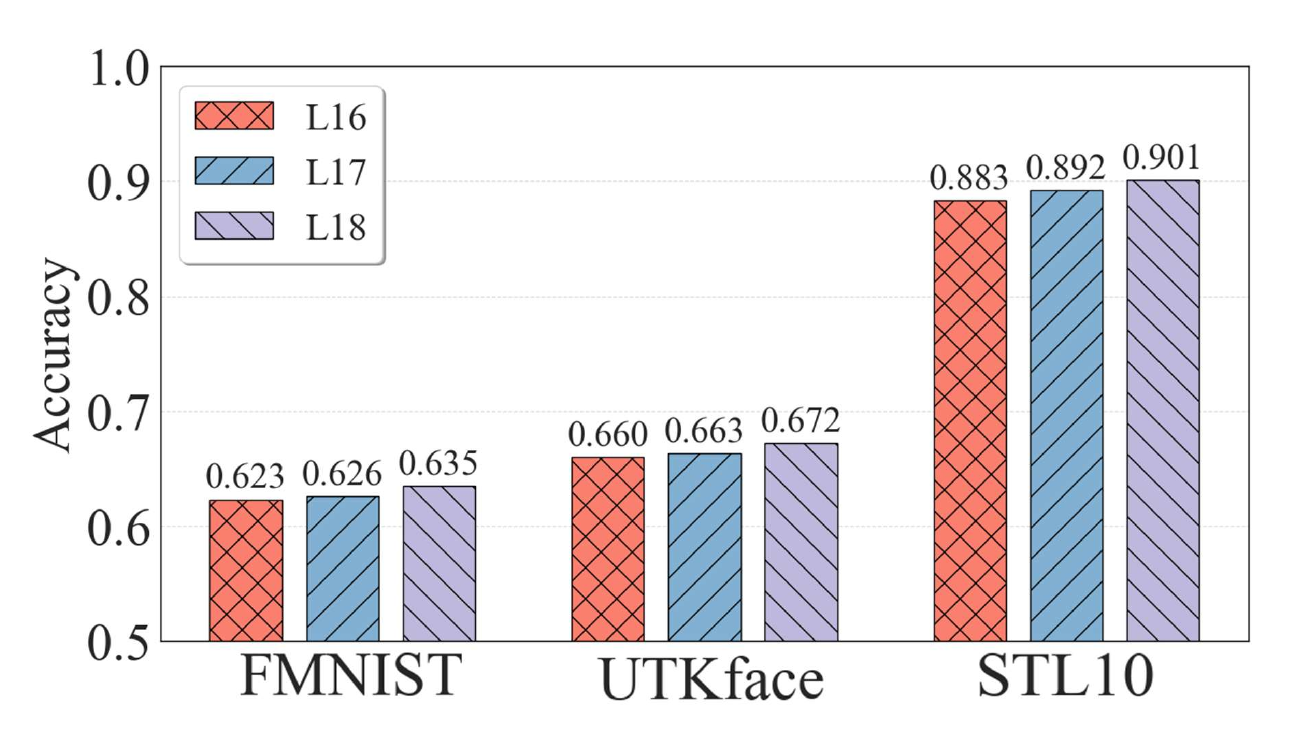}
        {ResNet}
    \end{minipage}%
    \caption{\small Attack accuracy trend when using 100\% of neurons from the last three layers.}
    \label{fig:Layerimpact}
    \vspace{-0.15in}
\end{figure}

In Table~\ref{tab:attackaccuracy} we report the attack accuracy and F1-score with the attack vector consisting of activations from every layer of the model, alongside the common attack features. The initial assessment indicates that activation values, without feature selection, do not lead to a stronger attack and can potentially even cause a weaker attack. We then assess the attack accuracy by employing a simple neuron selection strategy where we select all the neurons of a given layer to conduct the attack. Figure~\ref{fig:Layerimpact}, shows that even a simple layer selection strategy leads to improvement in attack accuracy. Notice how the deeper layers lead to higher attack accuracy. This is primarily because earlier layers learn general features and deeper layers learn task-specific features~\cite{YosClunBen14}. While overfitting is often attributed as the main cause for the success of MIA~\cite{liu2021ml}, for both UTFace and STL10 datasets we notice that the attack is more successful on ResNet18 than AlexNet even though ResNet18 has a lower overfitting rate. The primary cause of this is that the ResNet models, even with a lower loss and bias, have a higher variance on non-member data samples~\cite{BalSheHit22}. This causes the ResNet models to have a higher privacy leakage for some datasets.



\begin{table}[t]
\centering
\caption{\small The test accuracy/F1 score of the whitebox MIA without neuron selection.}
\label{tab:attackaccuracy}
\begin{tabularx}{\textwidth}{>{\centering\arraybackslash}p{2.5cm} >{\centering\arraybackslash}p{3cm} *{3}{>{\centering\arraybackslash}X}}
\toprule
\multirow{2}{*}{\makecell{Architecture}} & \multirow{2}{*}{\makecell{Approach}} & \multicolumn{3}{c}{Dataset} \\
\cmidrule(lr){3-5}
& & FMNIST & UTKFace & STL10 \\
\midrule
\multirow{2}{*}{AlexNet} & Ours w/o neuron selection & \multirow{2}{*} {0.554/0.670}& \multirow{2}{*}{0.619/0.679} &\multirow{2}{*}{0.775/0.802}  \\
\cmidrule{2-5}
& ML Doctor~\cite{liu2021ml} &0.570/0.680 &0.660/0.750 &0.820/0.840 \\
\midrule
\multirow{2}{*}{ResNet18} & Ours w/o neuron selection &\multirow{2}{*}{0.617/0.650 }& \multirow{2}{*}{0.691/0.721} &\multirow{2}{*}{0.915/0.927}  \\
\cmidrule{2-5}
& ML Doctor~\cite{liu2021ml} &0.550/0.560 &0.740/0.780 &0.905/0.880 \\
\bottomrule
\end{tabularx}
\vspace{-0.2in}
\end{table}

%

%
As discussed in Section~\ref{sec:methodology}, we further extend our layer selection strategy to identify the neurons with the highest contribution to develop more robust and efficient attacks. We then utilized the five mentioned statistical methods to rank the neurons based on the significant difference between these distributions and selected the top 20\%, 40\%, 60\%, and 80\%. We then train an attack model for each combination of neuron percentage and statistical technique, resulting in 20 attack models.

%
\begin{figure}[t]
    \centering
    \includegraphics[width=\textwidth]{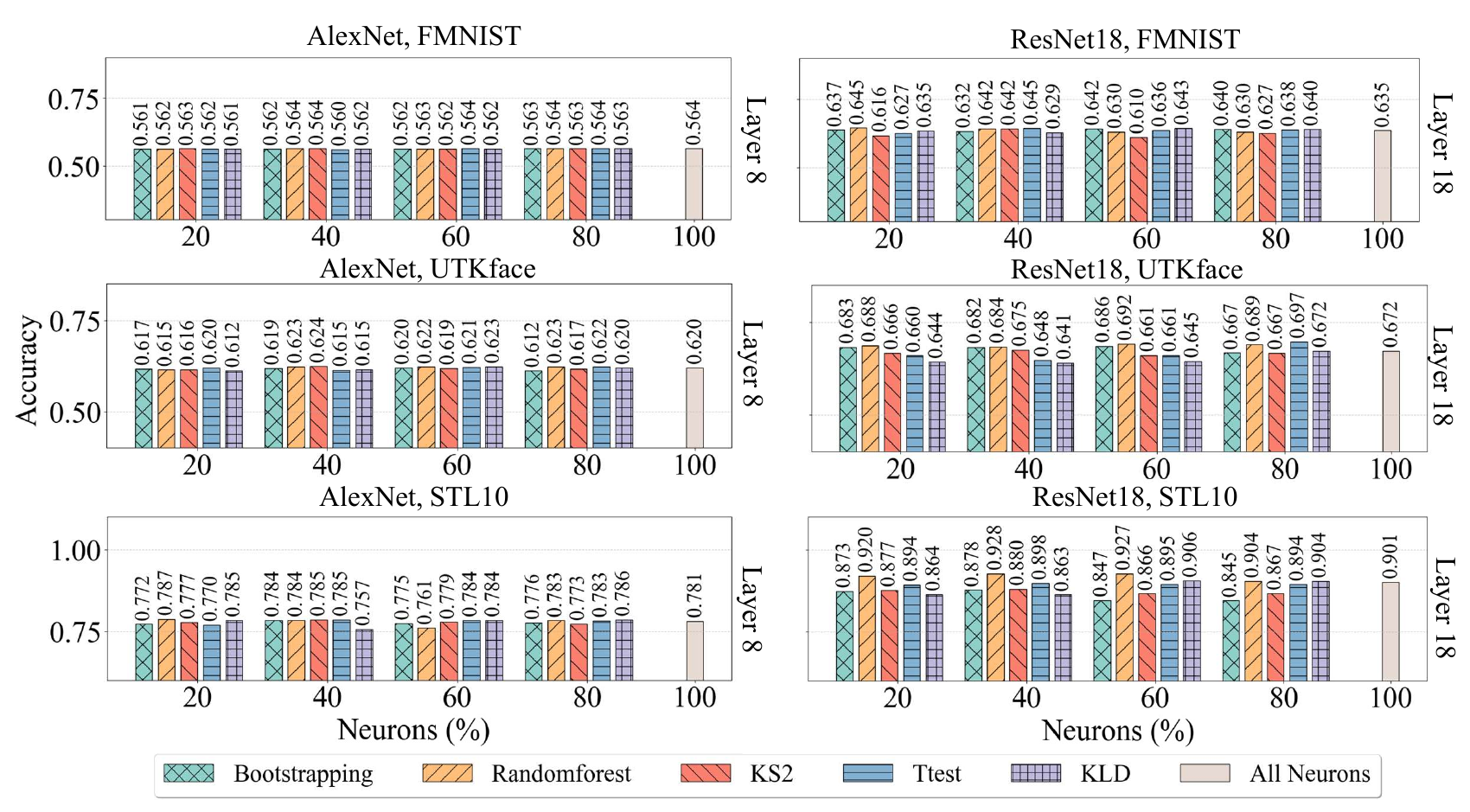} 
    \caption{\small MIA accuracy under different percentages of neurons of the last layer on two target models using five techniques across different datasets.}
    \label{fig:attack_acc_lastlayer}
    \vspace{-0.2in}
\end{figure}

Figure~\ref{fig:attack_acc_lastlayer} illustrates the accuracy of these 20 MIA attack models. We use the attack accuracy utilizing 100\% activations of the last layer as a baseline. For completeness, we also included the individualized attack accuracy of the other two layers in Figure~\ref{fig:combined_figure} in Appendix~\ref{appendix}. 
Across all dataset and architecture combinations, using just 20\% of the neurons to conduct the attack, irrespective of the selection technique,  leads to attack accuracies comparable to the baseline. This indicates that the majority of the activations do not contribute to privacy leakage. The results also show that random forest is the best-performing neuron selection strategy and using 40\% neurons with random forest, across all combinations, leads to the highest attack accuracy. We also notice that for some combinations increasing the percentage after 40\% of neurons can lead to a degradation in attack accuracy (e.g. ResNet18 trained on STL10). This primarily happens due to an increase in irrelevant activations, hence making the attack vector noisier.

\begin{figure}[!t]
    \centering
    \begin{subfigure}[b]{\textwidth}
        \centering
        \includegraphics[width=\textwidth]{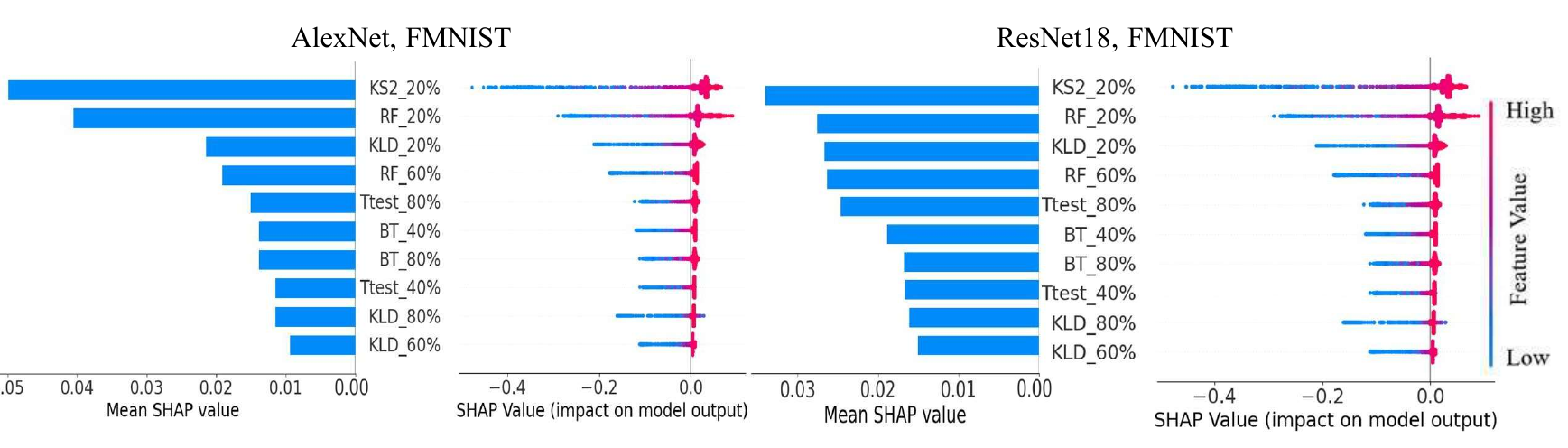}
    \end{subfigure}
    \begin{subfigure}[b]{\textwidth}
        \centering
        \includegraphics[width=\textwidth]{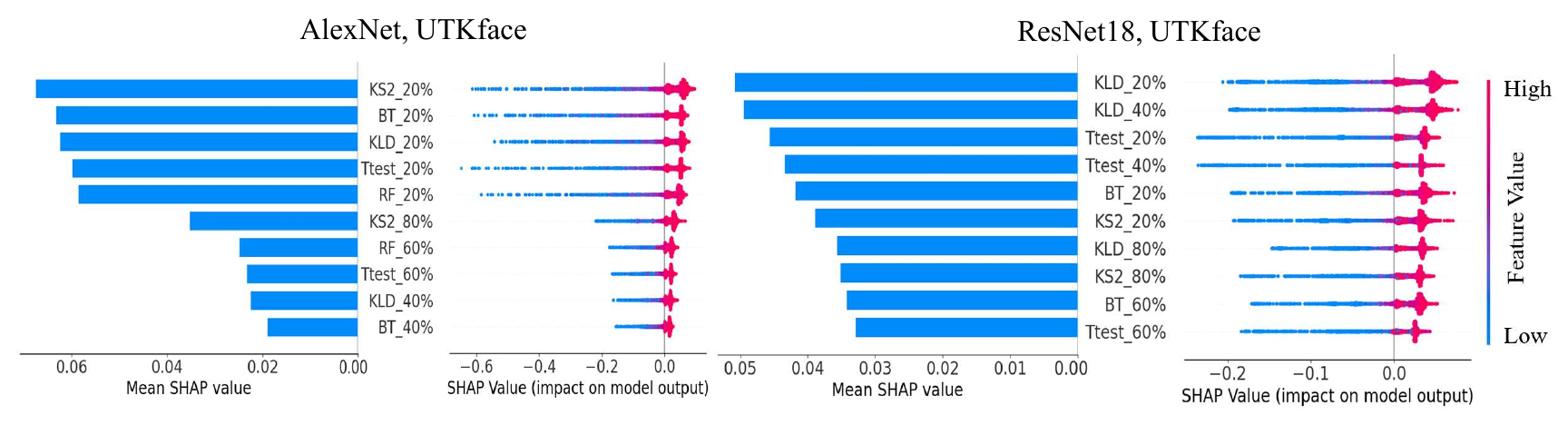}
    \end{subfigure}
    \begin{subfigure}[b]{\textwidth}
        \centering
        \includegraphics[width=\textwidth]{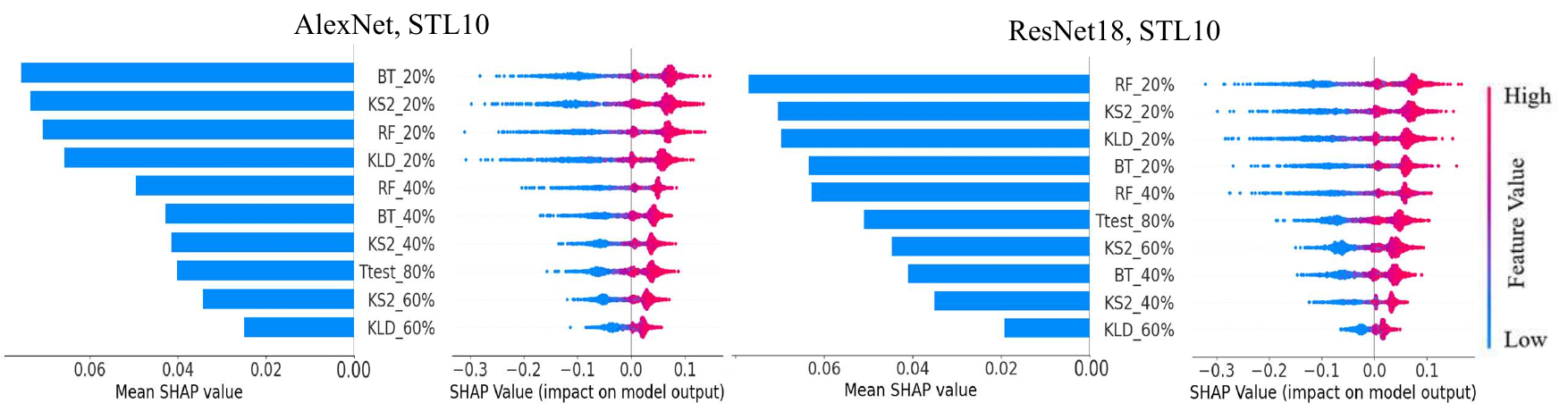}
    \end{subfigure}
    
    \caption{\small Comparative SHAP visualizations for membership inference attacks on AlexNet and ResNet18, highlighting the impact of varying neuron proportions in the last layer across FMNIST, UTKFace, and STL10 datasets, organized by rows.}
    \label{fig:shap}
    \vspace{-0.2in}
\end{figure}

\subsubsection{Feature Importance and Directional Impact}
We utilized SHAP to investigate the significance of neuron activations and to determine whether selecting a subset of neurons enhances MIA success compared to utilizing all neurons, and if so, to what extent. In particular, we use the mean SHAP values to rank the features by their importance. We also conducted a directional impact analysis on these features to identify the direction (\ie positive or negative) and the magnitude of their effect on the attack (Figure~\ref{fig:shap}). Note that we only visualize the top ten features due to space limitations.

The left bar plots in Figure~\ref{fig:shap} depict the impact of each of the top ten features on the attack's prediction accuracy. 
This analysis unveils a consistent trend across all datasets and architectures: neurons identified within the top 20\% and 40\%, as determined by various statistical methods, emerge as the primary contributing features for MIA success, with the 20\% exhibiting greater influence. This observation aligns with the previous evaluation, indicating that only a small subset of neurons carries more significant information regarding the membership status of data samples. We will use this analysis in our ensembling approach.

The right beeswarm plots visualize the distribution of SHAP values for each feature, \ie a particular set of neurons selected using one of the five statistical methods. Individual points represent the SHAP values per data sample, indicating the magnitude and direction of each feature's contribution to membership prediction. In this plot, the color indicates the magnitude of the feature values, normalized within the dataset; red and blue represent the high and low values, respectively. 

From Figure~\ref{fig:shap}, two trends were identified. Firstly, higher magnitudes of the selected neurons positively correlate with our whitebox MIA success, as evidenced by the higher density of red points with positive SHAP values. Conversely, samples with negative SHAP values (blue points) indicate that lower values of the selected neurons adversely impact MIA accuracy.
Another notable trend is the correlation between the high density regions of the distributions and attack accuracy. Specifically, our analysis shows that the STL10 dataset exhibits a distribution with a higher density at approximately 0.8 SHAP value when considering neurons with higher values (depicted in red), which is directly associated with the superior MIA accuracy observed for STL10 across all datasets. In contrast, the SHAP analysis for FMNIST results in a distribution with the smallest density distribution, at approximately 0.5), indicating a lower MIA accuracy.

\begin{figure}[t]
    \centering
    \includegraphics[width=\textwidth]{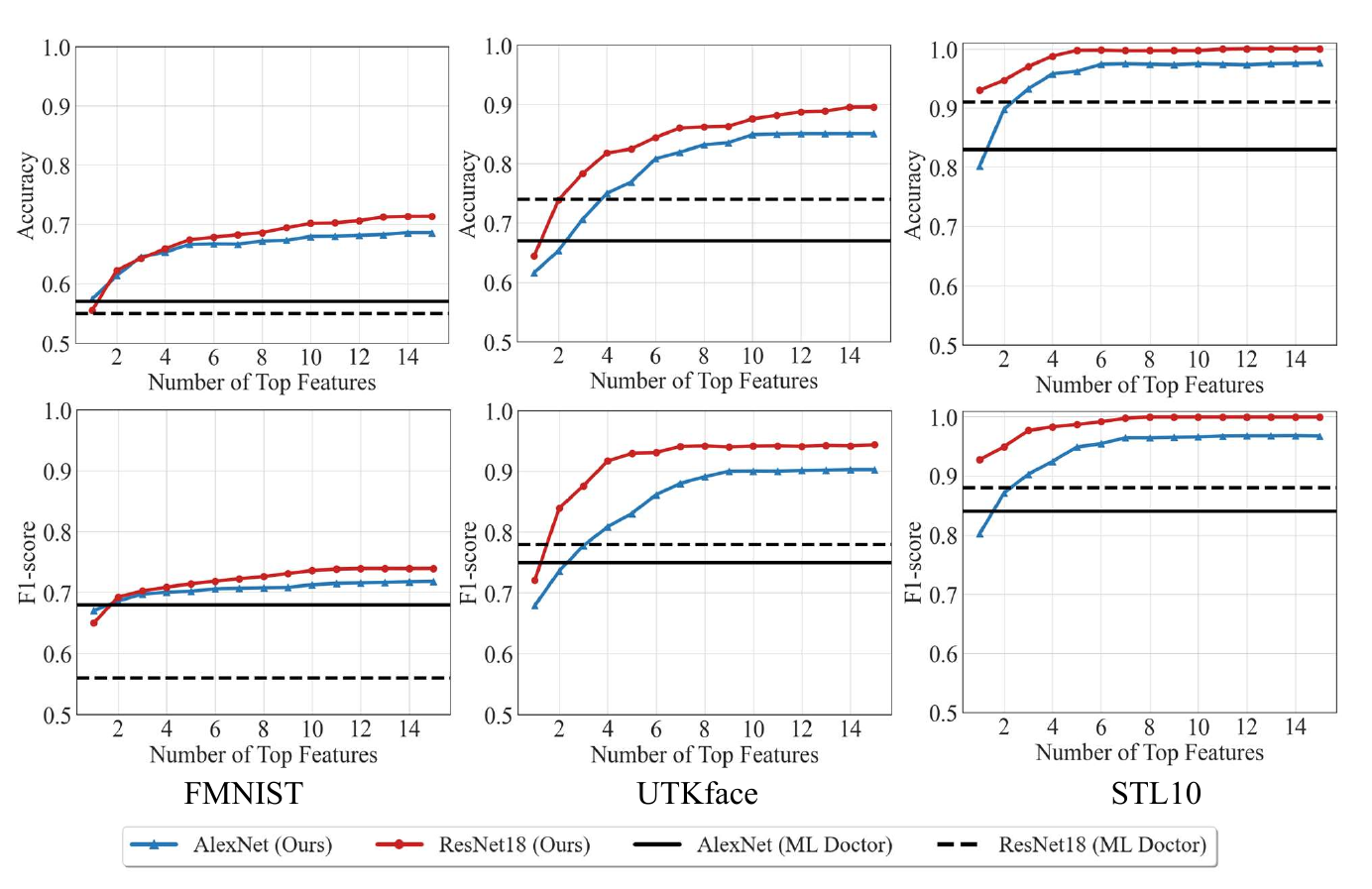}
        \vspace{-0.2in}
    \caption{\small (a) The first row illustrates the trend of the target test attack accuracy using target data when ensembling an increasing number of features. (b) The second row illustrates the trend of the target test attack F1-score using target data when ensembling an increasing number of features.}
    \label{fig:ensemble1}
    \vspace{-0.25in}
\end{figure}

\subsubsection{The Effect of Ensembling}
Using the feature importance analysis outcomes, we employed a stacked ensemble model to enhance our MIA robustness and accuracy. In particular, we ensemble the top-ranked features, which are heavily dependant on the dataset architecture combination, and assess it across an increasing number of features. 
%
%
Figure~\ref{fig:ensemble1} shows the accuracy and F1-score of the attack on the target model using target data with an increasing number of features for all dataset and architecture combinations.
It is evident that adopting a stacked ensemble approach significantly improves MIA accuracy across all datasets and architectures and can lead to an improvement up to 26.9\%. The consistent increase in the F1-score also indicates a noticeable decrease in false positive rates. Moreover, we observe that the overall MIA performance stabilizes after ensembling the top eight features, showcasing robustness towards increasing features. This shows a significant improvement over the state-of-the-art white-box MIA while also minimizing the attack vector. These findings show that an ensemble of statistical neuron selection techniques leads to a robust and scalable white-box MIA.

\subsection{Explainability Analysis of Raw Data Features}
\begin{figure}[t]
    \centering
    \includegraphics[width=\textwidth]{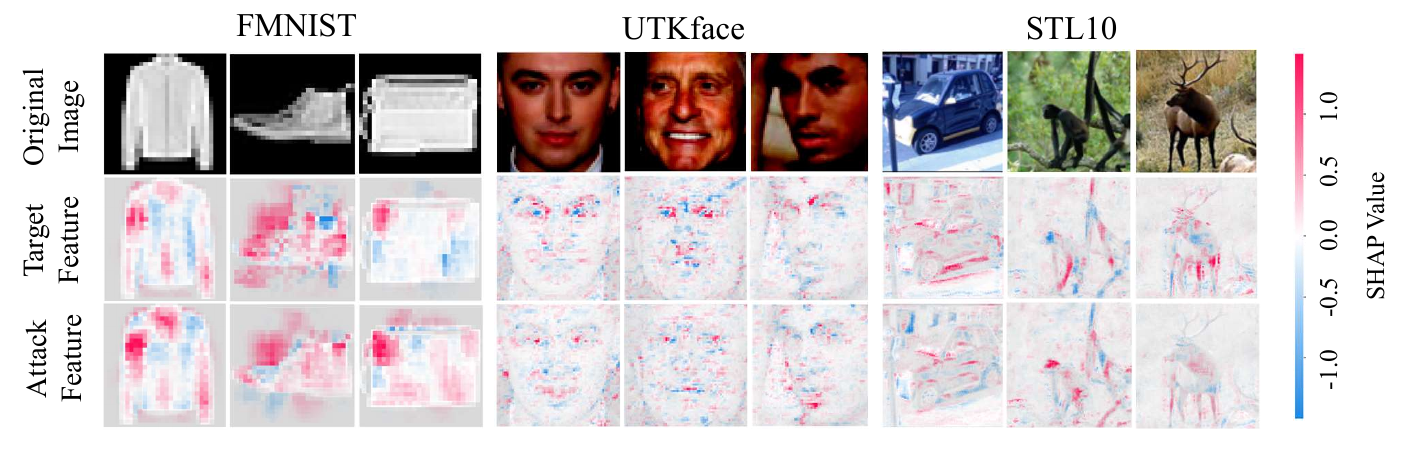}
    \vspace{-0.25in}
    \caption{\small SHAP visual analysis for the target and attack models on AlexNet with FMNIST, UTKface, and STL10 datasets.}
    \label{fig:FMNIST_acc}
    \vspace{-0.2in}
\end{figure}

In this subsection, we aim to achieve two objectives. Firstly, we seek to identify those raw features of the input data (\eg images) that are pivotal for the success of the target classification task, and those crucial for the MIA success. Subsequently, we aim to quantify the intersection between these feature sets.
Figure~\ref{fig:FMNIST_acc} visualizes the raw features crucial for the target classification task (second row) and the MIA (third row). It's important to note that the most influential features for the correct target/membership classification are those with higher SHAP values, highlighted in red. 
Our analysis indicates that contrasting the standalone target model with the combined target and attack models yields differing explanations for certain features, while some raw features are important for both tasks, the target classification and MIA, the visualization reveals additional features specific to attack prediction. For instance, when shown an image of a jacket (first column), the target model focuses on the shoulder and sleeve to classify it as a jacket. The attack model, however, also considers the collar region for inferring the membership status of this sample. However, these differences can be subtle and not human perceptible.
\begin{figure}[t]
    \centering
    \begin{minipage}[b]{0.5\textwidth}
        \centering
        \includegraphics[width=\textwidth]{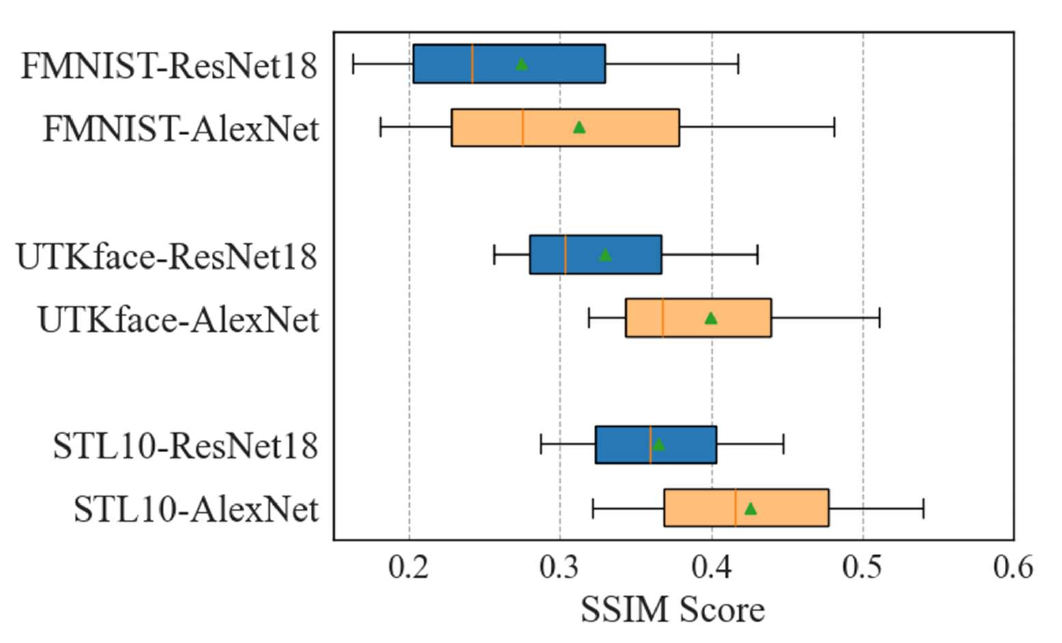}
        \caption*{(a) Member Data}
    \end{minipage}%
    \hfill 
    \begin{minipage}[b]{0.5\textwidth}
        \centering
        \includegraphics[width=\textwidth]{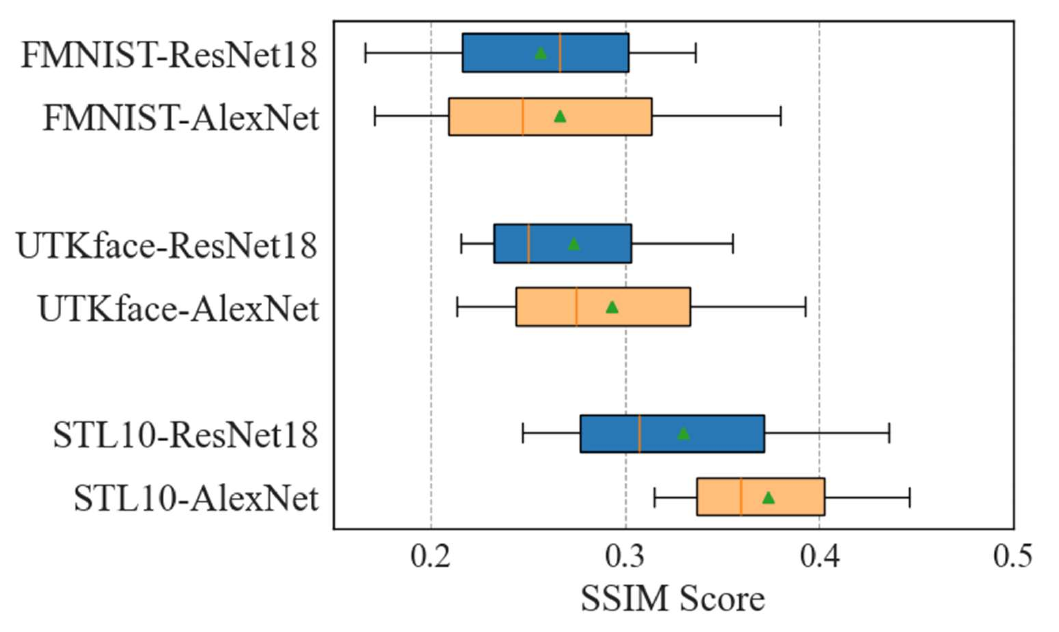}
        \caption*{(b) Non-member Data}
    \end{minipage}
    \caption{\small SSIM Score Comparison Across Multiple Datasets in AlexNet and ResNet18.}
    \label{fig:ssim}
    \vspace{-0.1in}
\end{figure}

To further quantify our analysis, we calculated the SSIM score between the SHAP output of the target model and that of the attack model. These experiments are extended across various datasets and model architectures, for both member and non-member samples. As illustrated in Figure~\ref{fig:ssim}(a), the mean SSIM value for the member data samples is capped at $0.42$, indicating a low overlap between the relevant features. Across different datasets, the SSIM score for FMNIST tends to be the lowest, followed by UTKface and STL10, consistent with the complexity of the datasets and the performance of the target classification task. These findings suggest that as the complexity of the dataset decreases, the differences become more pronounced.
We also observed that the SSIM scores when using AlexNet are marginally higher than those for ResNet18. This trend suggests that a higher MIA accuracy also leads to a lower overlap between the features impacting the target prediction and features impacting the attack. This also may be indicative of a correlation with neural network complexity, where more complex neural networks have less overlap between the feature maps.

%
We also conducted this analysis for the non-member data samples (as shown in Figure~\ref{fig:ssim}(b)) and observed that the same trends persisted across all datasets and architectures.

%% file: 7-Appendix.tex
\section{Appendix}
\label{appendix}

\begin{table}[ht]
\centering
\caption{Hyperparameters for Target, Shadow, Attack and SHAP Models.}
\label{tab:hyperparameters}
\begin{tabular}{lccccc}
\toprule
Model & Learning Rate & Optimizer & Epochs & Mini-batch Size  \\
\midrule
Target & $1 \times 10^{-5}$ & Adam & 300 & 64  \\
Shadow & $1 \times 10^{-5}$ & Adam & 300 & 64  \\
Attack & $1 \times 10^{-5}$ & Adam & 50  & 64  \\
SHAP & $1 \times 10^{-3}$ & Adam & 10 & 32  \\

\bottomrule
\end{tabular}
\end{table}

\begin{table}[h]
\centering
\caption{Attack model Architecture}
\label{tab:Attack}
\begin{tabular}{|p{4cm}|p{8cm}|}
\hline
\textbf{Name} & \textbf{Configuration} \\
\hline
Intermediate Component & 2 FC Layers (128, 64), ReLU, Dropout: 0.2 \\
\hline
Output Component & 2 FC Layers (128, 64), ReLU, Dropout: 0.2 \\
\hline
Loss Component & 2 FC Layers (1, 128, 64), ReLU, Dropout: 0.2 \\
\hline
Gradient Component & 1 Conv (5x5, Padding: 2), 1 BN, 1 MaxPool (2x2), 2 FC Layers (256, 128, 64), ReLU, Dropout: 0.2 \\
\hline
Label Component & 2 FC Layers (128, 64), ReLU, Dropout: 0.2 \\
\hline
Encoder Component & 4 FC Layers (320, 256, 128, 64, 2), ReLU, Dropout: 0.2 \\
\hline
\end{tabular}
\end{table}



%
\begin{figure}[p]
    \centering
    \begin{subfigure}[b]{\textwidth}
        \centering
        \includegraphics[width=\textwidth]{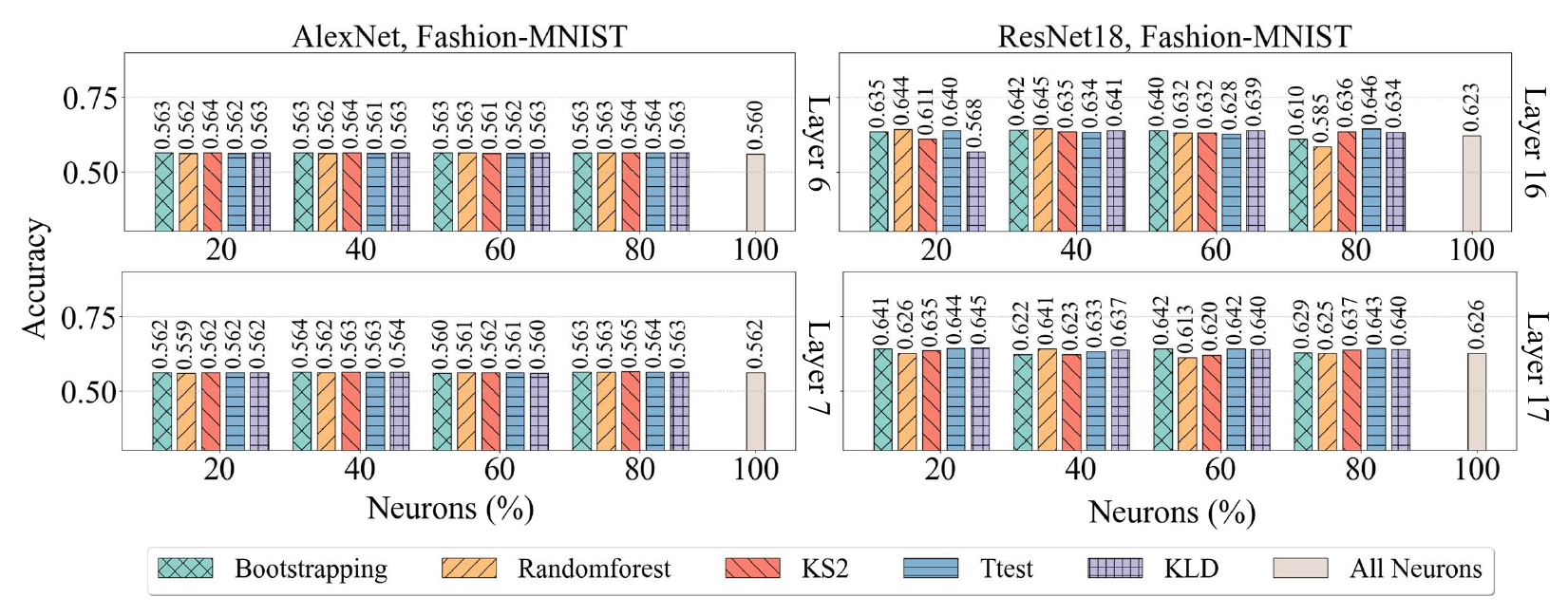}
    \end{subfigure}
    \begin{subfigure}[b]{\textwidth}
        \centering
        \includegraphics[width=\textwidth]{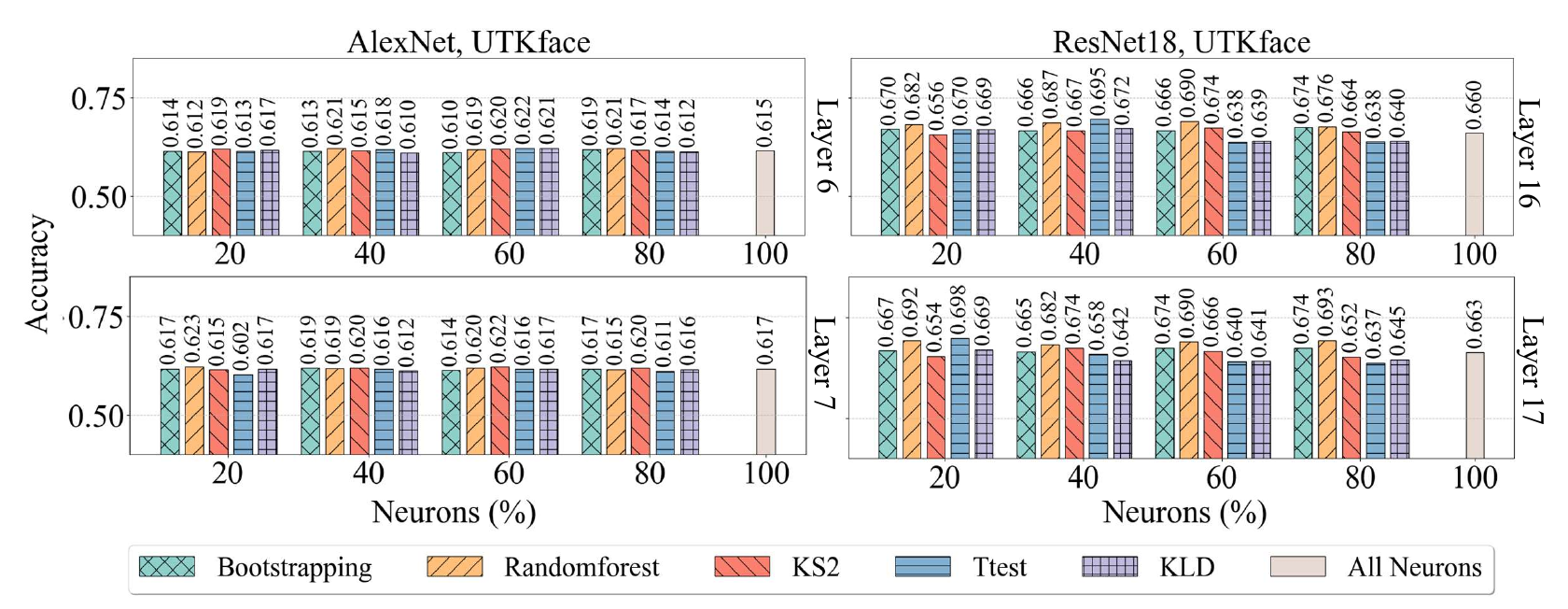}
    \end{subfigure}
    \begin{subfigure}[b]{\textwidth}
        \centering
        \includegraphics[width=\textwidth]{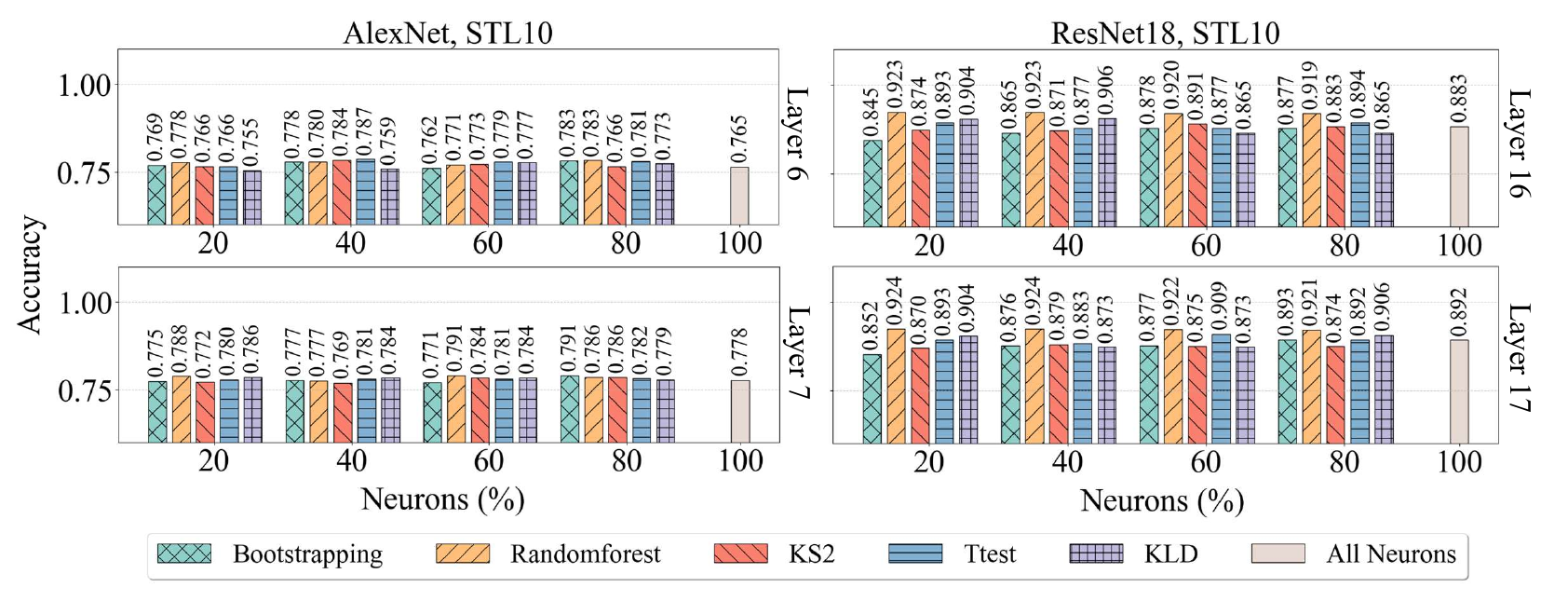}
    \end{subfigure}
    
    \caption{Accuracy of membership inference attacks under different percentages of neurons, different distinct analytical methods, and different target model architectures (left is AlexNet and right is ResNet18) on different dataset. }
    \label{fig:combined_figure}
\end{figure}